\documentclass{jfm}
\usepackage{graphicx}
\usepackage{epstopdf, epsfig}
\usepackage{amsmath}

\usepackage{xcolor}

\shorttitle{Dynamics generator GAN in fluid dynamics}
\shortauthor{Abdolvahhab. Rostamijavanani, Shanwu Li and Yongchao. Yang}

\title{Data-driven Modeling of Parameterized Nonlinear Fluid Dynamical Systems with a Dynamics-embedded Conditional Generative Adversarial Network}

\author{Abdolvahhab Rostamijavanani\aff{1}\corresp{\email{arostami@mtu.edu}},
  Shanwu Li\aff{1}
 \and Yongchao Yang\aff{1}}

\affiliation{\aff{1}Department of Mechanical Engineering-Engineering Mechanics, Michigan Technological University, Houghton, MI 49931, USA
}

\begin{document}

\maketitle

\begin{abstract}
   This work presents a data-driven solution to accurately predict parameterized nonlinear fluid dynamical systems using a dynamics-generator conditional GAN (Dyn-cGAN) as a surrogate model. The Dyn-cGAN includes a dynamics block within a modified conditional GAN, enabling the simultaneous identification of temporal dynamics and their dependence on system parameters. The learned Dyn-cGAN model takes into account the system parameters to predict the flow fields of the system accurately. We evaluate the effectiveness and limitations of the developed Dyn-cGAN through numerical studies of various parameterized nonlinear fluid dynamical systems, including flow over a cylinder and a 2-D cavity problem, with different Reynolds numbers. Furthermore, we examine how Reynolds number affects the accuracy of the predictions for both case studies. Additionally, we investigate the impact of the number of time steps involved in the process of dynamics block training on the accuracy of predictions, and we find that an optimal value exists based on errors and mutual information relative to the ground truth.  
\end{abstract}


\section{Introduction}
Computational Fluid Dynamics (CFD) is a field that uses numerical methods such as finite difference and finite volume to analyze fluid flow. By solving partial differential equations, such as the Navier Stokes equations, CFD aims to obtain accurate results through an iterative process. CFD offers several advantages over experimental methods. It provides a high level of fidelity, making it a reliable and cost-effective approach for fluid flow analysis. Additionally, CFD methods can handle different boundary conditions in a flexible manner, providing a more comprehensive picture of fluid flow.
However, the process of performing CFD simulations is resource-intensive and time-consuming, as a significant amount of computational power is required. This can make it difficult to explore the design space iteratively and evaluate different options quickly. The high cost of simulation and resource requirements make it challenging to perform CFD simulations in a timely and cost-effective manner (\cite{chen2020flowgan,geneva2020multi}).\par
 Data-driven surrogate models can be a useful tool for estimating the dynamics of parameterized nonlinear fluid dynamical systems, especially when physics knowledge is insufficient to develop closed-form analyses of dynamical behavior while measurements are available. In the past few years, machine learning techniques have been successfully applied to develop data-driven methods to model and identify nonlinear dynamics in a number of specific areas, including the discovery of governing equations (\cite{brunton2016discovering, quade2018sparse, kaheman2020sindy, cai2022online, li2019discovering}), linear representation of nonlinear dynamics (\cite{lusch2018deep}), nonlinear modal identification (\cite{li2021data,dervilis2019nonlinear,worden2017machine}), reduced-order models (\cite{li2021hierarchical,simpson2021machine}), and application in engineering (\cite{Li2018,huang2017bayesian,li2018condition,bao2017identification}). However, these methods have been intended for the specific application of \textit{fixed-parameter} nonlinear systems only; while data-driven modeling of \textit{parameterized} nonlinear systems is particularly challenging, primarily because the temporal dynamics as well as their dependence on certain parameters must be identified only from measurement data. Data-driven approaches have been presented for accelerating CFD simulations by employing deep neural networks (DNNs)~(\cite{bhatnagar2019prediction,guo2016convolutional}) which are capable of predicting the outcome of simulations directly. The DNN-based approach entails initially learning more complex features based on data produced by a full-order CFD solver as part of a learning process~(\cite{chen2020flowgan}). Using the learned model, one can predict flow fields for unknown flow problems depending on inputs that include flow conditions and initial conditions. By using a surrogate model instead of the numerous computation iterations that a CFD solver requires, predictive modeling is able to drastically reduce the time required for the generation of flow data (\cite{chen2020flowgan}). Recent developments in flow field prediction focus primarily on the prediction under fixed flow conditions~(\cite{bhatnagar2019prediction,wang2020towards}). A flow parameter such as the Reynolds number is typically used to quantify the flow conditions. During the design phase, flow conditions are subject to frequent changes. Therefore, a model that is trained for fixed parameters becomes obsolete if flow conditions change.

 Data-driven approaches, particularly those employing deep learning models, have shown promise in addressing the challenges of computational cost and scalability in CFD simulations. For instance, generative adversarial networks (GANs) have been utilized to accelerate fluid dynamic modeling by reconstructing high-dimensional, unsteady flow fields while maintaining accuracy and reducing computation time. Similarly, advancements in integrating energy-efficient and renewable resource-driven systems into smart building designs demonstrate the potential of machine learning for optimizing environmental and operational factors in complex systems (\cite{shafa6smart}). These efforts emphasize the role of intelligent systems in overcoming traditional limitations, thereby enhancing efficiency and sustainability (\cite{shafa6ranking}).
Models based on generative deep learning, including variational autoencoders (VAEs)~(\cite{kingma2013auto}) and generative adversarial networks (GANs)~(\cite{goodfellow2020generative}) offer great potential for developing a data-driven surrogate model of parameterized nonlinear fluid dynamical systems primarily due to their architectures: generating data based on fundamental parameters. Essentially, this process is quite similar to that used in conventional numerical simulations of parameterized systems, in which system responses are derived from the initial conditions and parameters of the system. A difference between the generative deep learning framework and the conventional numerical simulation is that the mathematical model mapping input parameters to output is unknown and should be identified only based on the data, i.e. a data-driven modeling problem; whereas for conventional numerical simulations, closed-form mathematical models (e.g., a set of parameterized governing equations) are available, which are derived from first principles with sufficient understanding of physics. Taking this basis into account, in this work, we utilize generative deep learning to develop a data-driven surrogate model to represent an unknown, yet measurable, parameterized nonlinear fluid dynamical system based on measurement data only.

Among a large number of generative models, GANs are distinguished by their distinctive advantages of the generative-adversarial architecture; this architecture employs a second model (discriminator) to adversarially enhance the generative modeling, notably in terms of learning the generative model. Recent work has been carried out on GANs and their variants as a tool for the data-driven modeling of parameterized nonlinear dynamical systems~(\cite{xie2018tempogan,farimani2017deep,strofer2018data,tsialiamanis2022application,ren2019fully,cheng2020data,ren2020learning,jegorova2020adversarial}). In particular, a conditional GAN has been devised for the simulation of incompressible flow and steady-state heat conduction~(\cite{farimani2017deep}); specifically, boundary conditions of the flow were treated as input information for the neural network. In addition, GANs have been used to enhance the spatial resolution of turbulent flow fields~(\cite{deng2019super}). It also has been used to reconstruct high-dimensional sequences of nonlinear unsteady flow fields~(\cite{cheng2020data}). There have been significant advancements in the application of generative deep learning in the context of identifying solid or fluid dynamics, yet several critical gaps remain. Most of the existing works do not directly describe the process of internal time evolution of dynamics because of the black-box mapping of input parameters (e.g., boundary conditions) to system states (e.g., fluid fields). Neither the dynamical state nor the mathematical model that describes the change in state over time are known. Therefore, these models may lack the ability to interpret physical phenomena and, therefore, are not suitable for generalization. It is therefore crucial to capture both the temporal dynamics and their dependence on parameters when developing data-driven surrogate models for parameterized nonlinear dynamical systems.

As a solution to this problem, this work presents a dynamics-generator conditional GAN (Dyn-cGAN) that can be used as a surrogate model for parameterized nonlinear fluid dynamical systems. Particularly, a dynamics block is incorporated within a modified conditional GAN, allowing identification of the temporal dynamics and their dependence on system parameters, concurrently. Taking into account the system parameters, the learned Dyn-cGAN model predicts the flow fields of the system in accordance with the parameters. We evaluate the developed Dyn-cGAN through numerical studies of different parameterized nonlinear fluid dynamical systems such as flow over a cylinder with different Reynolds numbers and a 2-D cavity problem with different sets of system parameters, and discuss the effectiveness and limitations of the model. 
\section{Problem formulation}
The Navier-Stokes~(N-S) equation is a set of partial differential equations that describe the motion and behavior of fluid substances. It models the conservation of mass, momentum, and energy, and accounts for the forces acting on a fluid such as pressure, viscosity, and external forces:
\begin{equation}\label{ganeq1}
\frac{\partial \boldsymbol{u}}{\partial t} + \boldsymbol{u} \cdot \nabla \boldsymbol{u} = -\frac{1}{\rho} \nabla p + \nu \nabla^2 \boldsymbol{u} + \boldsymbol{f}
\end{equation}

where $\boldsymbol{u}$ is the velocity field, $\rho$ is the fluid density, $p$ is the pressure, $\nu$ is the fluid kinematic viscosity, and $\boldsymbol{f}$ is a source term. We examine fluid flows with nonlinear dynamics that are parametrized by a given number of simulation parameters~($d$)~$P_{sim} \in \rm{\Omega} \subset \mathbb{R}^{d}$, whereby the parameter domain of interest~$\rm{\Omega}$ is usually bounded:
\begin{equation}\label{eq:s1}
\begin{aligned}
    &{\mathbf{s}}(t;P_{sim}) = \mathcal{N}\left(\mathbf{s}(t=0;P_{sim}),P_{sim}\right)\\
    &\text{with} \quad \mathbf{s}(t=0;P_{sim})=\mathbf{s}^{0}
\end{aligned}
\end{equation}
where $\mathbf{s}(t;P_{sim}) \in \mathbb{R}^{n}$ is the flow field~(velocity or pressure), $\mathbf{s}^{0}$ represents the initial condition, and $\mathcal{N}(\cdot)$ stands for the nonlinear function which represents the underlying nonlinear dynamics of a fluid flow. Solving the governing equation \eqref{eq:s1} provides future flow fields~(future velocity or pressure fields) over a specified time steps ($T$) of a fluid flow given initial conditions and system parameters:
\begin{equation}\label{eq:s2}
    \mathbf{s}(t_{1},\ldots,t_{T};P_{sim}) = {\mathcal{F}}\left(\mathbf{s}^{0},P_{sim}\right)
\end{equation}
where ${\mathcal{F}}$ refers to the solution of the PDE (equation.\eqref{eq:s1}). 

For systems whose governing equation (equation.\eqref{eq:s1}) may be unknown because of the lack of physics knowledge required for first-principle modeling, the solution function ${\mathcal{F}}$ can be approximated by a surrogate model which is generally more efficient than numerically solving the original N-S equation and can be identified from measurement data only. To achieve this while retaining the physical interpretation of the model for understanding the transient dynamics, this work presents a Dyn-cGAN surrogate model to approximate and identify the solution function ${\mathcal{F}}$ conditioned on the flow parameters, using a \textit{dynamics} block to capture the state transition in the latent coordinates ($\varphi$):
\begin{equation}\label{eq33}
 \varphi^{t+1}=\mathcal{A}\left(\varphi^{t}\right)  
\end{equation}\label{eq3}
 and reconstruct it to the original flow fields~($\mathbf{s}$) as shown in Fig. \ref{FIG:2}. $\mathcal{A}$ is the nonlinear transformation function to encapsulate state transition identified and approximated by the dynamics block of the Dyn-Gen model, and $\varphi^{t}$ refers to the latent coordinates generated by Dyn-cGAN at time $t$. This recursive prediction may be carried out for a specified prediction time steps $T$.

\begin{figure}
	\centering
	\includegraphics[width=0.6\textwidth]{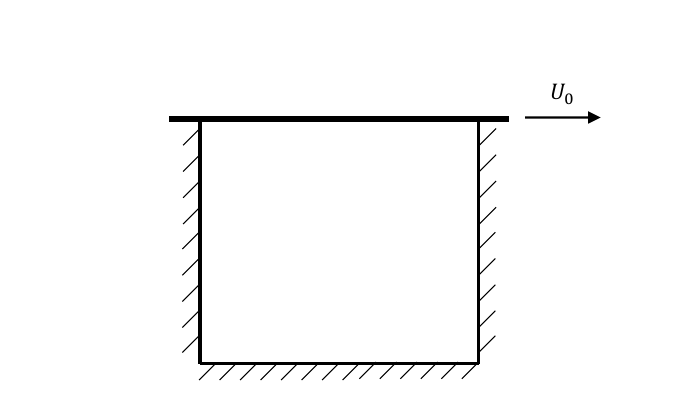}
	\caption{two-dimensional cavity problem: flow has a rich fluid flow physics including  multiple rotating
recirculating regions specially on the corners of the cavity which is dependent on the Reynolds number.   }
	\label{FIG:1}
\end{figure}

\section{Data-driven modeling of parameterized nonlinear fluid dynamical systems}
Our presented framework offers a distinctive capability in fluid dynamics prediction, which is the ability to predict the temporal fluid behavior based on the Reynolds number of the stream. This sets our framework apart from other methods by providing a more comprehensive and accurate prediction of fluid behavior, as it takes into account the key characteristic of fluid flow, the Reynolds number. By incorporating this crucial factor, the framework can provide a more accurate representation of the fluid's behavior, including steady-state, and transitional flow. 
\subsection{Objective}
The aim of this work is to present a dynamics-generator conditional GAN (Dyn-cGAN) specialized in modeling and identifying parameterized fluid flows based on data only. As a unique feature of this network, a dynamics block is incorporated into the generator model in order to recursively predict the evolution of flow fields over multiple time steps in relation to system (physical) parameters. There are two components to the presented Dyn-cGAN: a dynamics-generator (Dyn-Gen) and a discriminator. Dyn-Gen learns to generate a sequence of flow fields according to a given set of system parameter values; while the discriminator contributes to the training process of Dyn-Gen through an adversarial process. The trained Dyn-Gen serves as the identified model of the parameterized nonlinear fluid dynamical systems that can be used to simulate or predict sequences of flow fields based on any possible system parameter values.
\subsection{Dynamics-generator (Dyn-Gen)}
\begin{figure}
	\centering
	\includegraphics[width=1\textwidth]{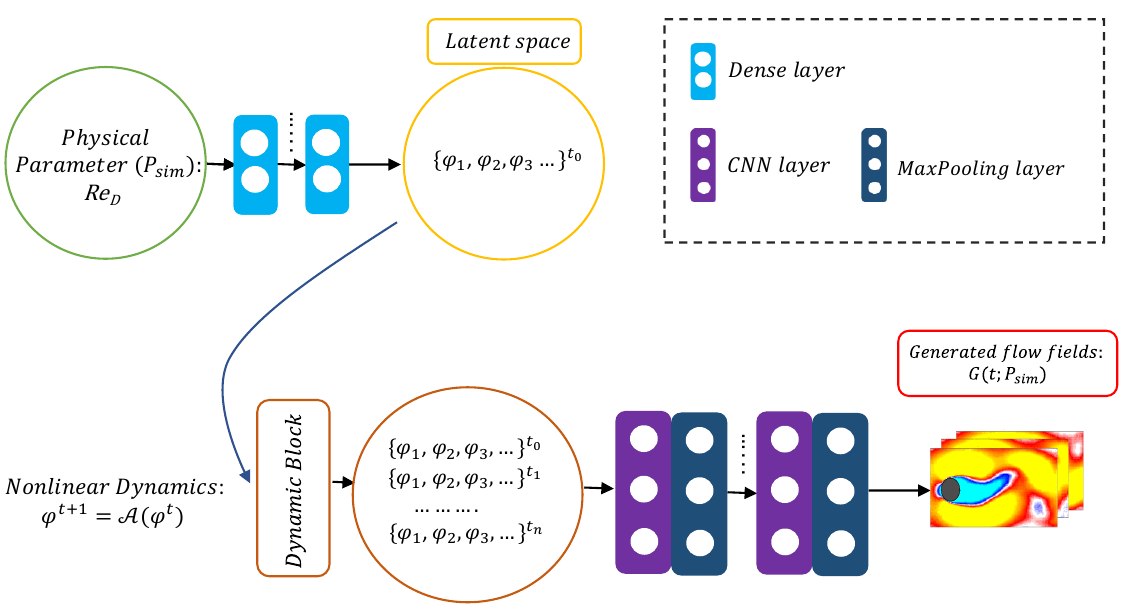}
	\caption{Dyn-Gen Model architecture: using few but important information of system, this model generates multi-step predicted flow fields with embedded dynamic block  }
	\label{FIG:2}
\end{figure}

The system (physical) parameters are inputs as vectors in the Dyn-Gen framework (see Fig. \ref{FIG:2}). By processing physical parameters through layers that are densely connected with nonlinear activation functions, a latent vector is produced that encodes the nonlinear dynamics behind the flow field. Therefore, the overall objective is to train the network to capture the intrinsic dynamical features of a flow based on the inputs (physical parameters) provided.\par
The latent vectors (that may represent some intrinsic coordinates of the system, such as modal coordinates) are then passed through to the dynamics block in order to predict the next time step values, and the predicted time step value is then used to predict the next time step, and the process is repeated for a specified number of time steps, which is a hyperparameter. Accordingly, the generator represents a physics-integrated model that can accurately capture the dynamics of the system given its system (physical) parameters.\par
Fig. \ref{FIG:2} illustrates the architecture of the Dyn-Gen.
The low-dimensional representation of the original flow field (latent space coordinates) which is predicted recursively in dynamics blocks is then decoded into physical coordinates (full flow field) through some convolutional and maxpooling layers. During this step, predicted flow fields are compared with ground truths. Using this approach, the Dyn-Gen produces flow fields as close to ground truth as possible. Moreover, the dynamics block meets the dynamic requirements of the system.\par
The Dyn-Gen model is trained using two loss functions. The first relates to prediction. Given a set of parameters, the objective is to predict flow fields as closely as possible to the ground truths. To quantify the error of prediction for a given time period, we can use the mean squared error (MSE) loss function. The loss function is described as follows:
\begin{equation}
   \lambda_{pred}^{G}= \frac{1}{n}\sum_{i=1}^{n} ||G(t;P^{i}_{sim}),\mathbf{s}(t;P^{i}_{sim})||_{MSE}
\end{equation}
where $G$ denotes generator model, $\mathbf{s}(t;P_{sim})$ is the ground truth flow fields, and $n$ is the number of training samples. The objective of this loss is to minimize the MSE between the predicted flow and the actual flow in terms of features. The MSE loss is considered \textit{hard} as it evaluates closeness from a deterministic prediction viewpoint. However, this hard requirement may result in challenges during training and a lack of robustness in the model obtained. \par
To address these issues, a \textit{soft} requirement is introduced, requiring the distribution of the generator-predicted flow fields to be similar to the ground truth. This is achieved by employing the generative-adversarial concept with the addition of a discriminator. The generator is then trained with an adversarial loss ($\lambda_{adv}^{G}$) to generate flow fields with a distribution that resembles the ground truth, thus attempting to deceive the discriminator:\par

\begin{equation}
   \lambda_{adv}^{G}= -\frac{1}{n}\sum_{i=1}^{n} log D(G(t;P^{i}_{sim}))
\end{equation}
where $D$ refers to the discriminator model. The minimum loss occurs when the discriminator cannot distinguish between the generated samples and the ground truths~(between generated probability distribution and corresponding real data probability distribution). By combining adversarial and prediction loss functions, the Dyn-Gen model is updated as follows:
\begin{equation}
   \lambda^{G}= \beta_{1}\lambda_{pred}^{G}+ \beta_{2}\lambda_{adv}^{G}
\end{equation}
where $\beta_{1}$ and $\beta_{1}$ are loss weights (hyper-parameters) of prediction and adversarial loss functions that need to be tuned over training process. Ultimately, by striking a balance between the hard requirement of the deterministic prediction ($\lambda_{pred}^{G}$) and the soft requirement of the probabilistic distribution ($\lambda_{adv}^{G}$), the generator is anticipated to undergo easier training and exhibit robustness. 

\subsubsection{Discriminator}
\begin{figure}
	\centering
	\includegraphics[width=1\textwidth]{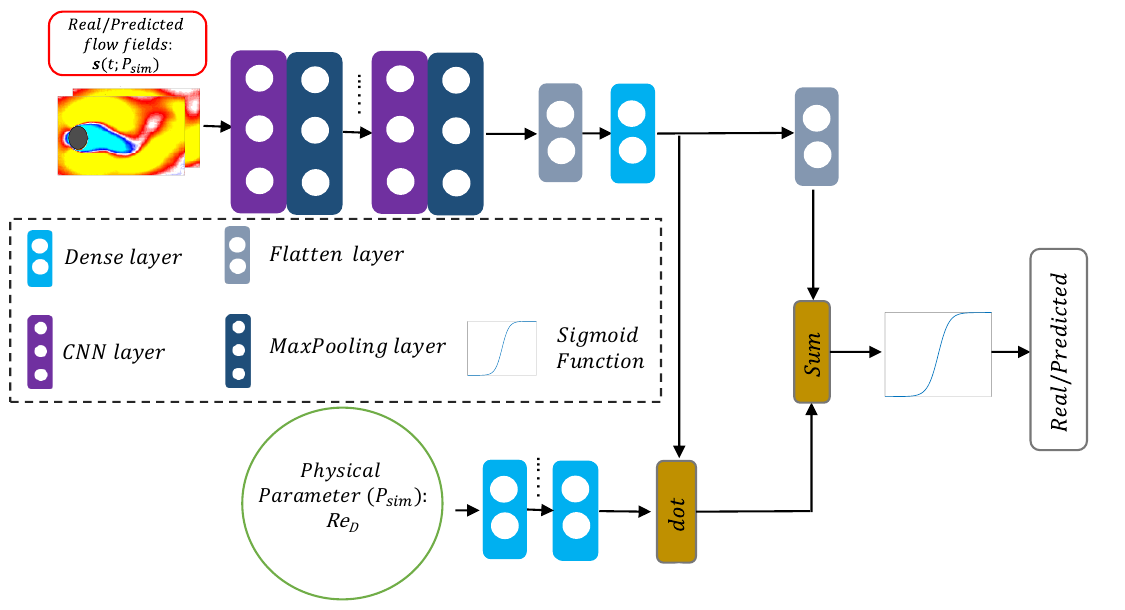}
	\caption{Discriminator Model: using fake/predicted flow fields with their corresponding physical parameters~($Re_D$), the discriminator distinguish the real data with the presented architecture   }
	\label{FIG:3}
\end{figure}

Described above, the discriminator is intended to assist in the training of the Dyn-Gen. Essentially, in the discriminator (see Fig. \ref{FIG:3}), flow fields generated by Dyn-Gen and real ones are fed into the discriminator on the one hand, and the physical parameters for each corresponding sample are also fed into the network on the other hand. As a result, we have a conditional GAN since the input data do not simply consist of pure noise; they contain relevant information about the dynamics of our system. During the process of model reduction, each 2-D flow field passes through some downsampling layers. As a result, a vector is produced after passing through these layers.\par
On the other hand, the physical parameters pass through some dense layers to produce a latent vector as we had in the Dyn-Gen model.
Now, the element-wise dot product of these two vectors (a vector derived from ground truth or predicted flow fields and another derived from physical parameters) can be achieved. As a result, information regarding fake~(predicted) and real samples is merged together with their respective parameters (output is a single value). After flow fields have been flattened and pushed through a dense layer, they produce a single value that will be added to the single value previously mentioned. By applying the sigmoid function, it generates a value between 0 and 1 that indicates whether the sample conditioned on the system parameters is real (ground truth) or fake~(predicted). Following is the adversarial loss function for the discriminator model:
\begin{equation}
    \lambda_{adv}^{D}= -\frac{\gamma}{n}\sum_{i=1}^{n} log D(G(t;P^{i}_{sim}))+log (1-D(\mathbf{s}(t;P^{i}_{sim})))
\end{equation}
where $\gamma$ is the hyperparameter~(loss weight) for adversarial loss which estimates the divergence between the distribution of the generated flow fields and the real ones. This results in an endless game between two models until an equilibrium situation is achieved if we consider the generator model as an attacker attempting to generate samples that fool the discriminator model as a defender.
\subsection{Network architecture and training}
The network architecture information is depicted in Fig. \ref{FIG:2} and Fig. \ref{FIG:3}. Due to the nonlinear dynamic nature of the systems studied, all activation functions used in layers are nonlinear. There are different nonlinear activation functions such as Relu, Sigmoid, and Tanh in Keras which is high-level API in Tensorflow. Relu is one the widely used functions which has faster training run time~(\cite{lusch2018deep}) which is used in our networks. It should be noted that the dynamics block is a multi-layer perception model and the last layer of dynamics block has linear activation functions. Adam optimizer with slow learning rate, $\alpha=0.00001$ is used for both Dyn-Gen and discriminator. In order to initialize the weights in each model, the Xavier initialization method~(\cite{glorot2010understanding}) is applied. There are hidden layers in the form of ${\textit{Wa + b}}$ that are followed by nonlinear activation function, in which $\textit{W}$ and $\textit{b}$ correspond to weights and biases, respectively, and $\textit{a}$ is referred to as input data. The Xavier initialization method generates a random number that is distributed uniformly along the range of $-\frac{1}{\sqrt{\mathbf{\eta}}}$ and $\frac{1}{\sqrt{\eta}}$, where $\eta$ represents the number of inputs to the node. We evaluate the performance of our Dyn-cGAN during a variety of training sessions~(hyperparameters-tuning). Several sets of hyperparameters is examined (weights of loss functions) and the results are based upon hyperparameters in which there is the least validation error (Table.~\ref{tab:1}).
\begin{table*}\label{tab:1}
	\caption{Loss weights for each case study}
	\label{tab:1}       
	\begin{tabular}{lp{2.5cm}p{2.5cm}p{2.5cm}}
		\hline\noalign{\smallskip}
	Case study  & $\beta_{1}$  & $\beta_{2}$  & $\gamma$ \\
		\noalign{\smallskip}\hline\noalign{\smallskip}
		Stream-wise velocity  & 100 & 10 & 10 \\
		Transverse velocity  & 100 & 10 & 10 \\
Stream-wise velocity (transient) & 100 & 10 & 10 \\
Transverse velocity (transient) & 100 & 1 & 10 \\
2-D Cavity Problem (Transverse velocity) & 100 & 1 & 10
 \\
 2-D Cavity Problem (vorticity) & 100 & 1 & 10 
 \\
		\noalign{\smallskip}\hline
	\end{tabular}
\end{table*}
\section{Results and Discussions}
To assess the efficiency and effectiveness of Dyn-cGAN, two case studies are performed: the flow over a cylinder in both transient and steady-state conditions, and the transient condition of a 2-D cavity problem~(\cite{peng2003transition,bruneau20062d}). The data sets for each case consist of 100 samples, each with 50 time steps, except otherwise stated. Each sample has a different Reynolds number while the channel size remains constant. Reynolds number is the key factor that determines the flow behavior in both case studies.\par
\subsection{Flow over a cylinder}
Our method is not restricted to any specific type of flow, but we focus on a two-dimensional flow over a cylinder, which is a commonly used example in previous studies (\cite{raissi2020hidden,maulik2020reduced}, to demonstrate our approach.\par 
The flow over a cylinder creates vortex shading in its wake, known as the Karman vortex street. The flow is in steady-state with Reynolds number ranging from $100<R_{D}<200$. The Navier-Stokes equation, described in equation \eqref{ganeq1}, governs the flow. A no-slip boundary condition is applied. The channel is divided into 96 by 192 evenly spaced grid points and 100 different test setups (each with a unique Reynolds number) are examined, with each setup containing 50 time steps. Both steady-state and transient conditions of the flow are analyzed for each velocity field~(stream-wise and transverse velocities).\par
\subsubsection{Steady-state condition}
In steady-state flow, the properties of a fluid such as velocity, pressure, and temperature remain constant over time at any given point within the system. This means that the flow of the fluid moves continuously without significant changes in its behavior as it moves through the system. The network is trained with the stream-wise and transverse velocity fields, separately, and different physical parameters are condensed into one key value, the Reynolds number. \par
\begin{figure}
	\centering
	\includegraphics[width=1\textwidth]{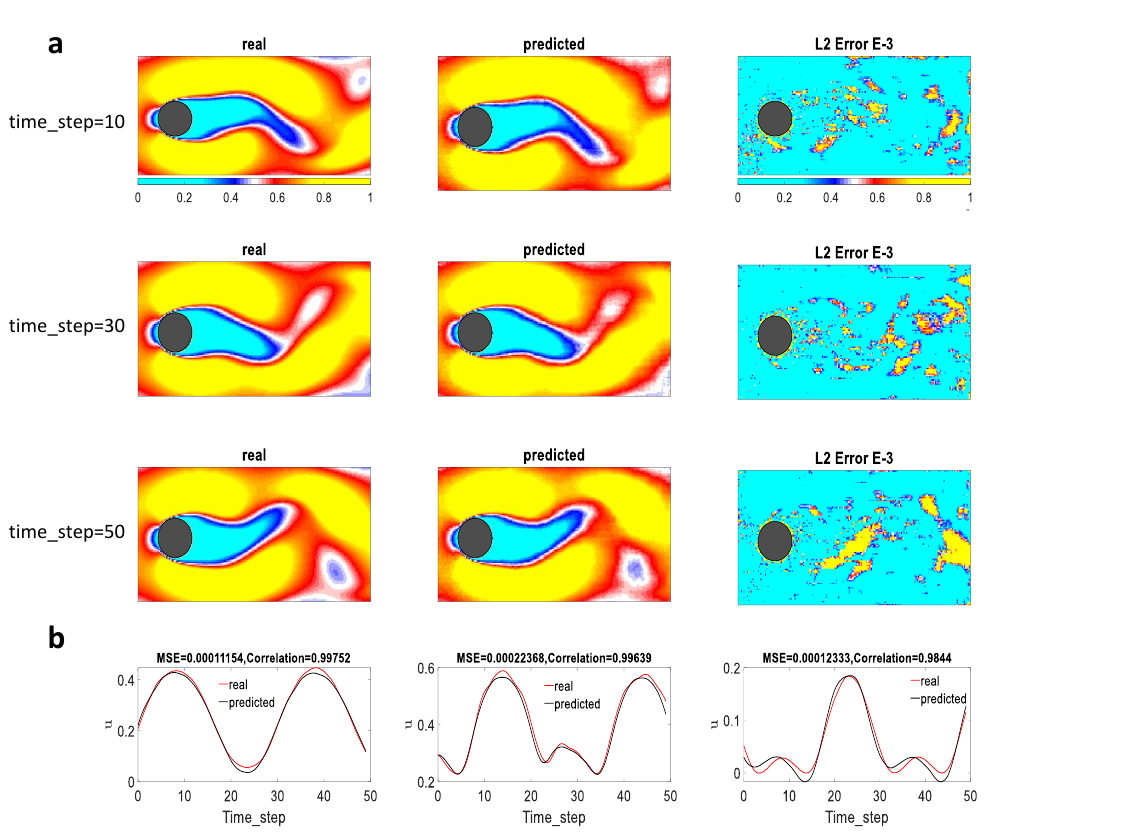}
	\caption{Steady-state stream-wise velocity  prediction for three time steps and corresponding $L2$ Error in spatial domain and time domain prediction for three randomly selected points in the flow}
	\label{FIG:4}
\end{figure}

Fig. \ref{FIG:4}(a) displays the spatial prediction of the stream-wise velocity field using Dyn-cGAN at three different time steps, along with the spatial L2 error for each time step. The prediction is seen to match the ground truth data well. Fig. \ref{FIG:4}(b) illustrates the effectiveness of the dynamics block by showing the trajectory of three randomly selected physical points in the spatial domain over time, spanning the entire 50-time step period. The MSE and correlation coefficient are used to quantify the performance.\par
\begin{figure}
	\centering
	\includegraphics[width=1\textwidth]{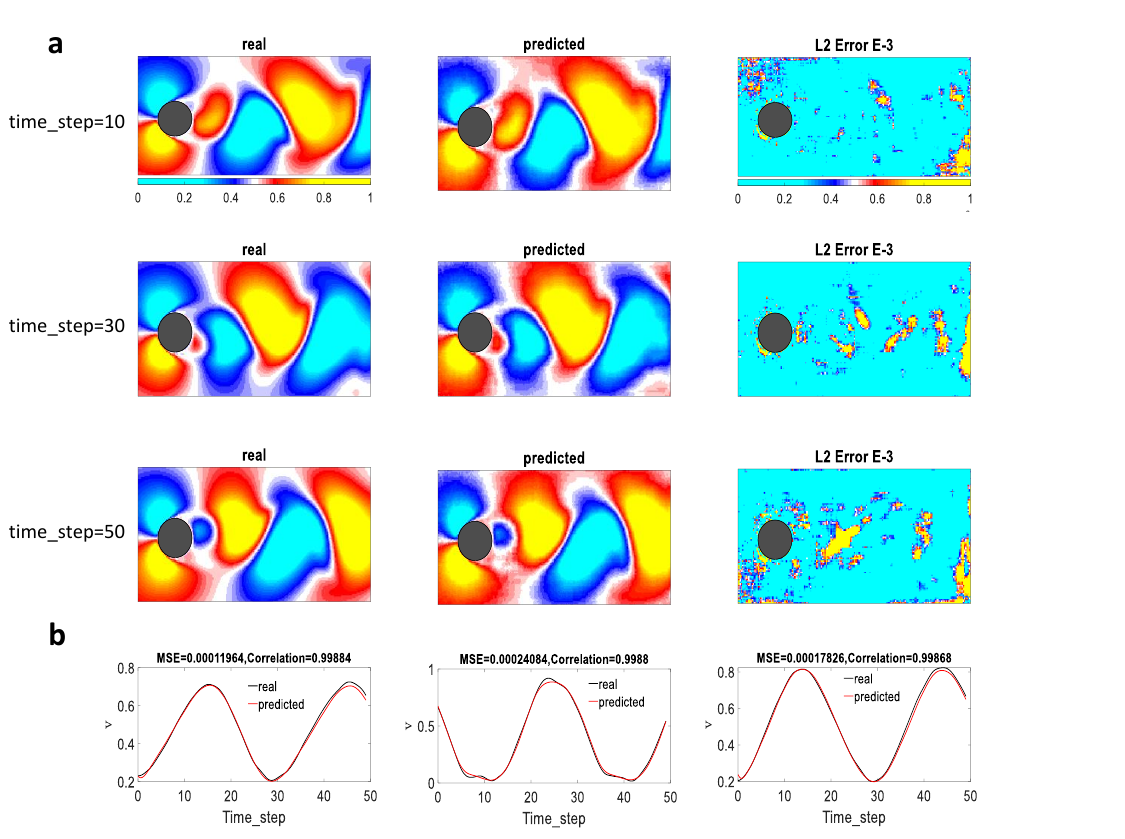}
	\caption{Steady-state transverse velocity  prediction for three time steps and corresponding $L2$ Error in spatial domain and time domain prediction for three randomly selected points in the flow}
	\label{FIG:5}
\end{figure}

A similar investigation is performed for the transverse velocity field. Fig. \ref{FIG:5}(a) shows the predicted fields at various time steps, with the full trajectories of three random physical points shown in Fig. \ref{FIG:5}(b). This figure also displays the L2 errors in the spatial domain and the MSE and correlation coefficient in the time domain.\par
Fig.~\ref{FIG:9} displays the streamlines of flow at a specific time step, demonstrating Dyn-cGAN's ability to estimate velocity fields in both directions. The predicted streamlines are in good agreement with the actual streamlines, as they have similar angles.\par
\begin{figure}
	\centering
	\includegraphics[width=1\textwidth]{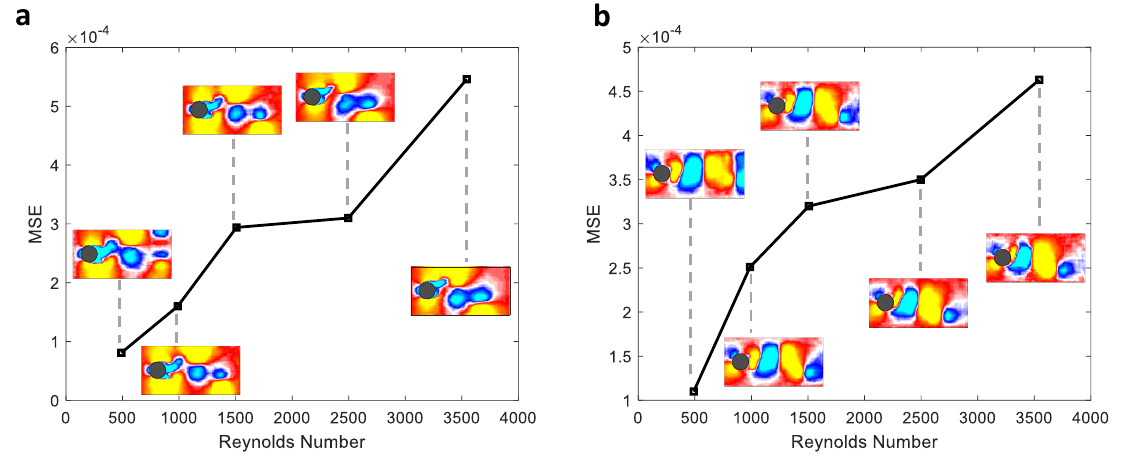}
	\caption{Effect of Reynolds number on the prediction of Dyn-cGAN. Two main plots~(\textbf{left}: stream-wise velocity and \textbf{right}: transverse velocity) display the flow fields of fluid at different Reynolds numbers for each stream. The Reynolds number is a measure of the balance between inertial and viscous forces in a fluid and is a crucial factor in fluid dynamics.}
	\label{FIG:6}
\end{figure}
\subsubsection{Influence of Reynolds Number on predictions}
The unique ability of our presented framework is the prediction of temporal fluid dynamics conditioned on the Reynolds number of the stream. In this section, the effect of Reynolds number on Dyn-cGAN predictions is analyzed. The training set features Reynolds numbers from 400 to 3500. The channel grid remains the same and each sample in the training set includes 50 time steps. The simulation includes stream-wise and transverse velocity data for flow over a cylinder. Fig. \ref{FIG:6} shows that the prediction accuracy decreases as Reynolds number increases due to the increasing complexity of flow making it more challenging for Dyn-cGAN to predict accurately.
\subsubsection{Transient condition}
\begin{figure}
	\centering
	\includegraphics[width=1\textwidth]{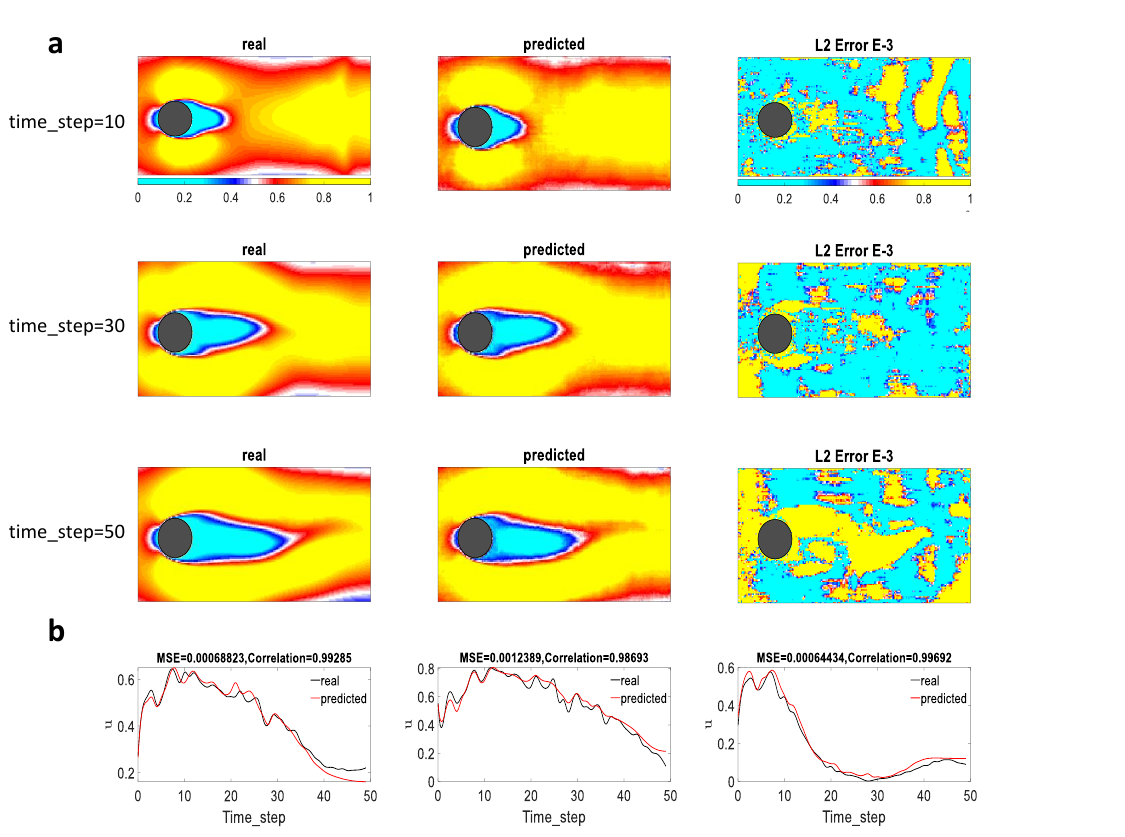}
	\caption{Transient stream-wise velocity  prediction for three time steps and corresponding $L2$ Error in spatial domain and time domain prediction for three randomly selected points in the flow}
	\label{FIG:7}
\end{figure}

\begin{figure}
	\centering
	\includegraphics[width=1\textwidth]{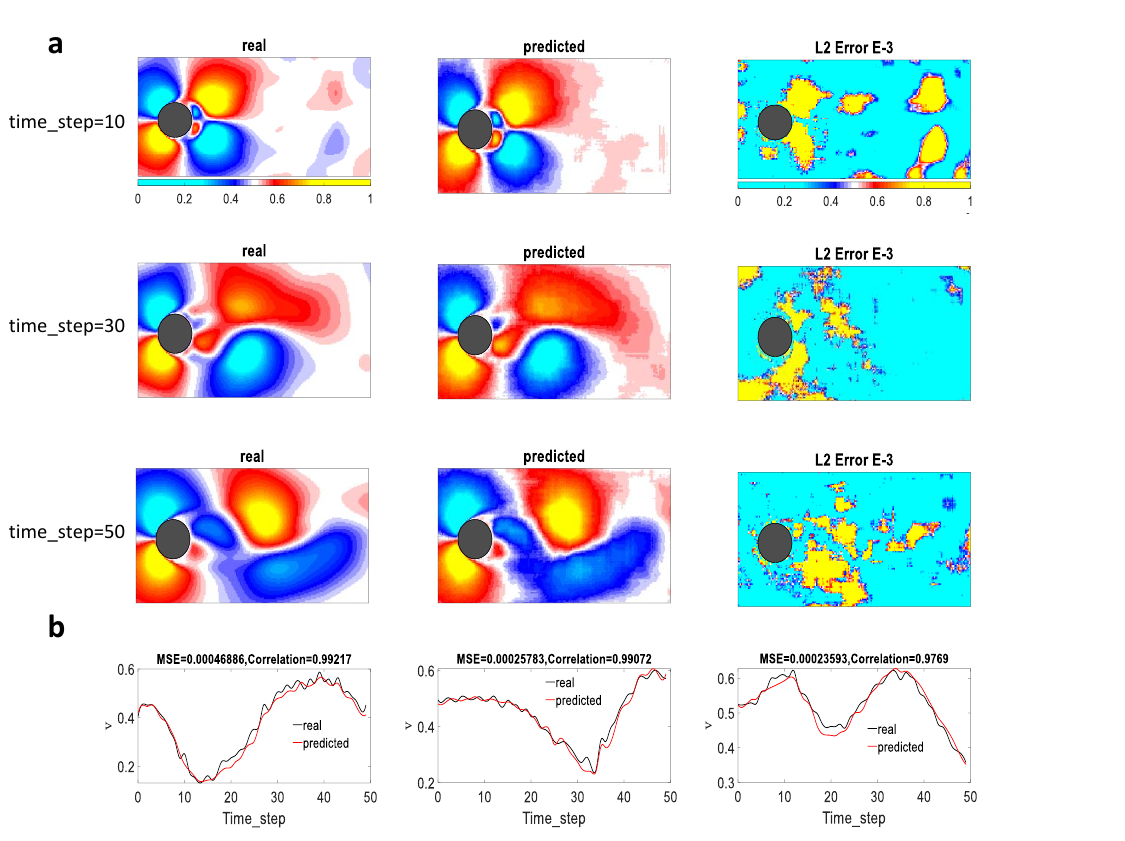}
	\caption{Transient transverse velocity  prediction for three time steps and corresponding $L2$ Error in spatial domain and time domain prediction for three randomly selected points in the flow}
	\label{FIG:8}
\end{figure}

Transient conditions refer to a fluid flow that is changing over time. This type of flow can be characterized by fluctuations in velocity and pressure, and is often seen in systems where there are disturbances to the flow, such as changes in the geometry of the system or changes in the fluid's velocity. In these situations, the fluid must adjust to the changes, which can result in complex and dynamic flow patterns.
This part explores the prediction of the transient flow conditions over a cylinder, both for the stream-wise and transverse velocity fields. The performance of the Dyn-cGAN in predicting the velocity fields both spatially and temporally is shown in Fig. \ref{FIG:7} and Fig. \ref{FIG:8} for stream-wise and transverse velocity, respectively. Transient conditions are generally more challenging for networks than steady-state conditions, as evidenced by the increase in spatial and temporal errors compared to steady-state conditions.\par

\subsection{2-D Cavity Problem}
\subsubsection{Transient transverse velocity}
Another example of 2-D flow is the transient flow inside a square cavity whose lid has a constant velocity (as depicted in Fig.~\ref{FIG:1}). This is governed by the incompressible Navier-Stokes equations derived in equation. \eqref{ganeq1}. The flow's transverse velocity inside the cavity is studied. The channel is evenly divided into 128 by 128 grids, with a Reynolds number range of 400 to 500.\par
\begin{figure}
	\centering
	\includegraphics[width=1\textwidth]{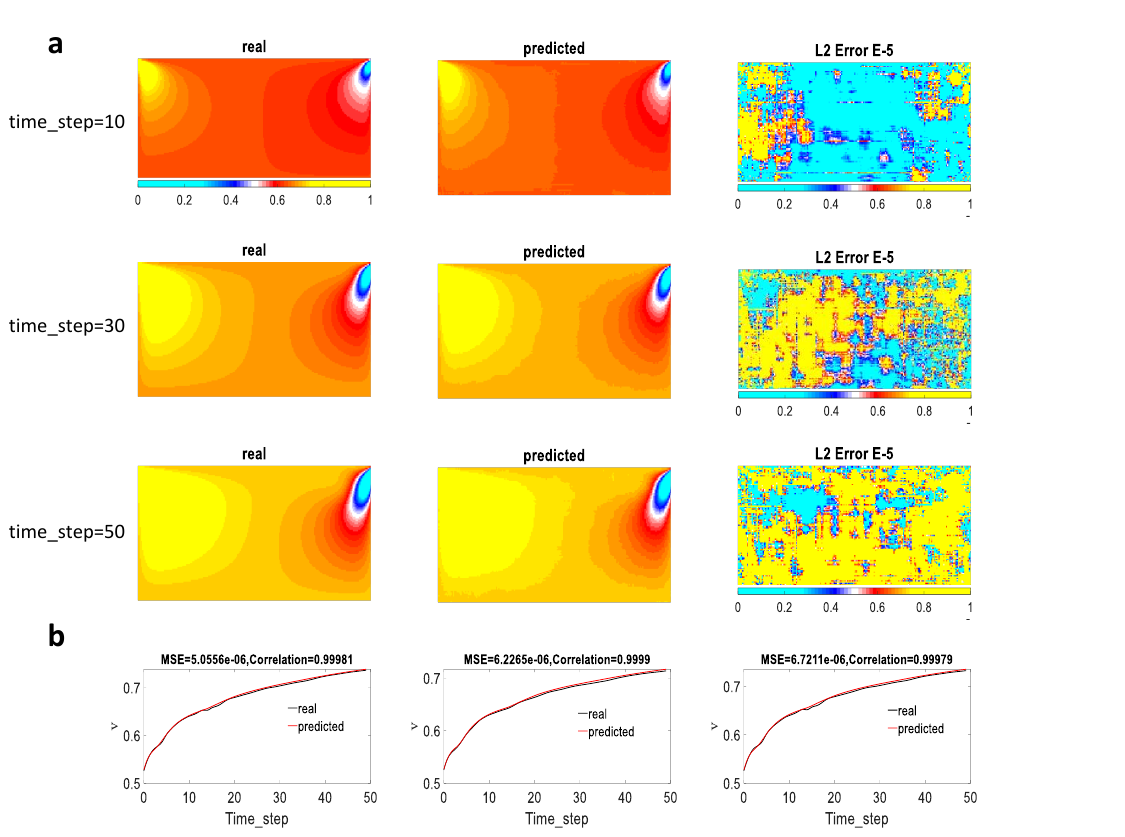}
	\caption{Transient transverse velocity  prediction of a 2-D cavity problem for three time steps and corresponding $L2$ Error in spatial domain and time domain prediction for three randomly selected points in the flow}
	\label{FIG:9}
\end{figure}

The transient transverse velocity of a cavity problem is investigated by passing a lid with constant velocity over the channel. As depicted in Fig.~\ref{FIG:9}, both the spatial and temporal predictions have very small errors, which indicates that the presented Dyn-cGAN is capable of handling different types of flow fields, including transient conditions.\par
\begin{figure}
	\centering
	\includegraphics[width=1\textwidth]{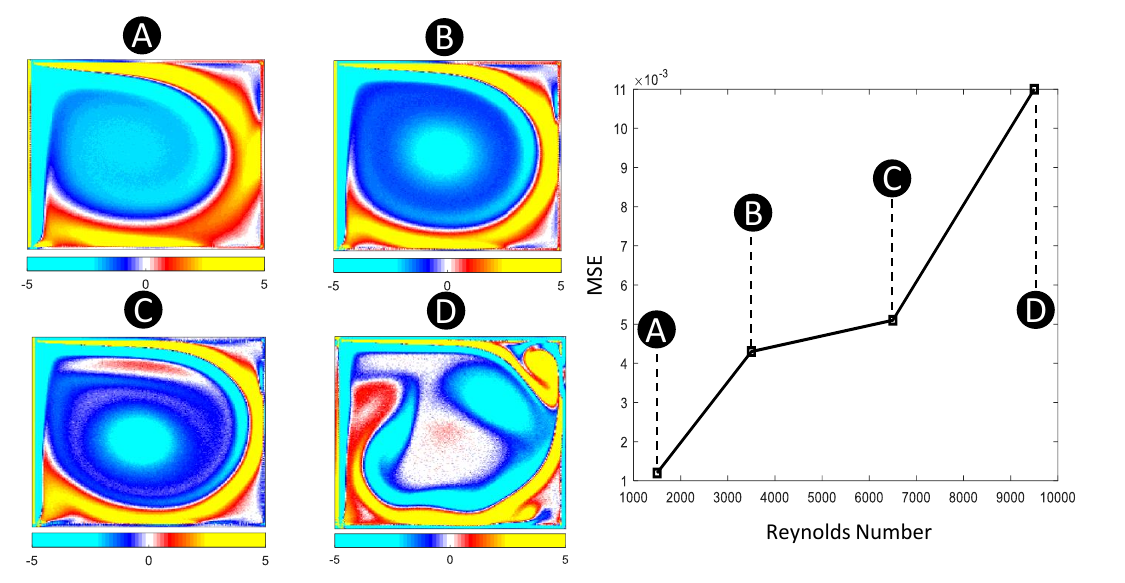}
	\caption{Effect of Reynolds number on the prediction of Dyn-cGAN for 2-D cavity vorticity}
	\label{cav}
\end{figure}
\subsubsection{Influence of Reynolds number: Transient vorticity}
 The aim of the transient vorticity lid-driven cavity problem is to comprehend the intricate movements of the fluid, which includes the formation of vortices, and other types of flow patterns as the top lid is driven in linear motion. This problem is considered transient because it involves changes in fluid properties over time, as opposed to steady-state flow where fluid properties are constant. The effect of Reynolds number on Dyn-cGAN's prediction for a 2D cavity problem is analyzed. The Reynolds number in the training data varies from 1000 to 10000. The channel is divided into 192 by 192 grids, with 30 time steps in each sample set. The simulations involve vorticity data. Fig. \ref{cav} shows a decrease in prediction accuracy with an increase in Reynolds number. As Reynolds number increases, the flow complexity increases, making it harder for Dyn-cGAN to make accurate predictions.
\subsection{Influence of the prediction horizon in the training}
The most notable feature of the presented Dyn-cGAN is the incorporation of dynamics into the generator, referred to as the Dyn-Gen. For training the Dyn-Gen, the required prediction horizon is one of the most critical hyperparameters. A short training horizon (a small number of time steps, $T$, in equation. \eqref{eq33}) facilitates convergence, however, the trained model is not robust to error accumulation during testing where a long sequence of flow fields is predicted recursively. On the other hand, training over a long period of time (a large number of time steps, $T$, in equation. \eqref{eq33}) yields greater robustness to error accumulation, although convergence may be difficult. Therefore, for the ultimate performance of the trained model, it is necessary to choose an (approximately) optimal prediction horizon that balances the training convergence and the model robustness.

Thus, we investigate the effect of the number of time steps (the prediction horizon) of the training phase on the testing accuracy of flow field predictions. Our model is trained with different numbers of time steps and the trained model's performance is tested for a specific number of time steps (200 time steps). To ensure a fair training condition for each scenario, we implement an early stop condition, a technique widely applied in deep learning frameworks. Our early stopping strategy involves two factors; one is the maximum number of training epochs, and the other is a tolerance, which indicates how patient we should be to allow training losses to decrease over time. Each training has a maximum of 15000 epochs, and we evaluate whether the training MSE loss value decreases over a period of 200 consecutive epochs before stopping it. 

\begin{figure}
	\centering
	\includegraphics[width=1\textwidth]{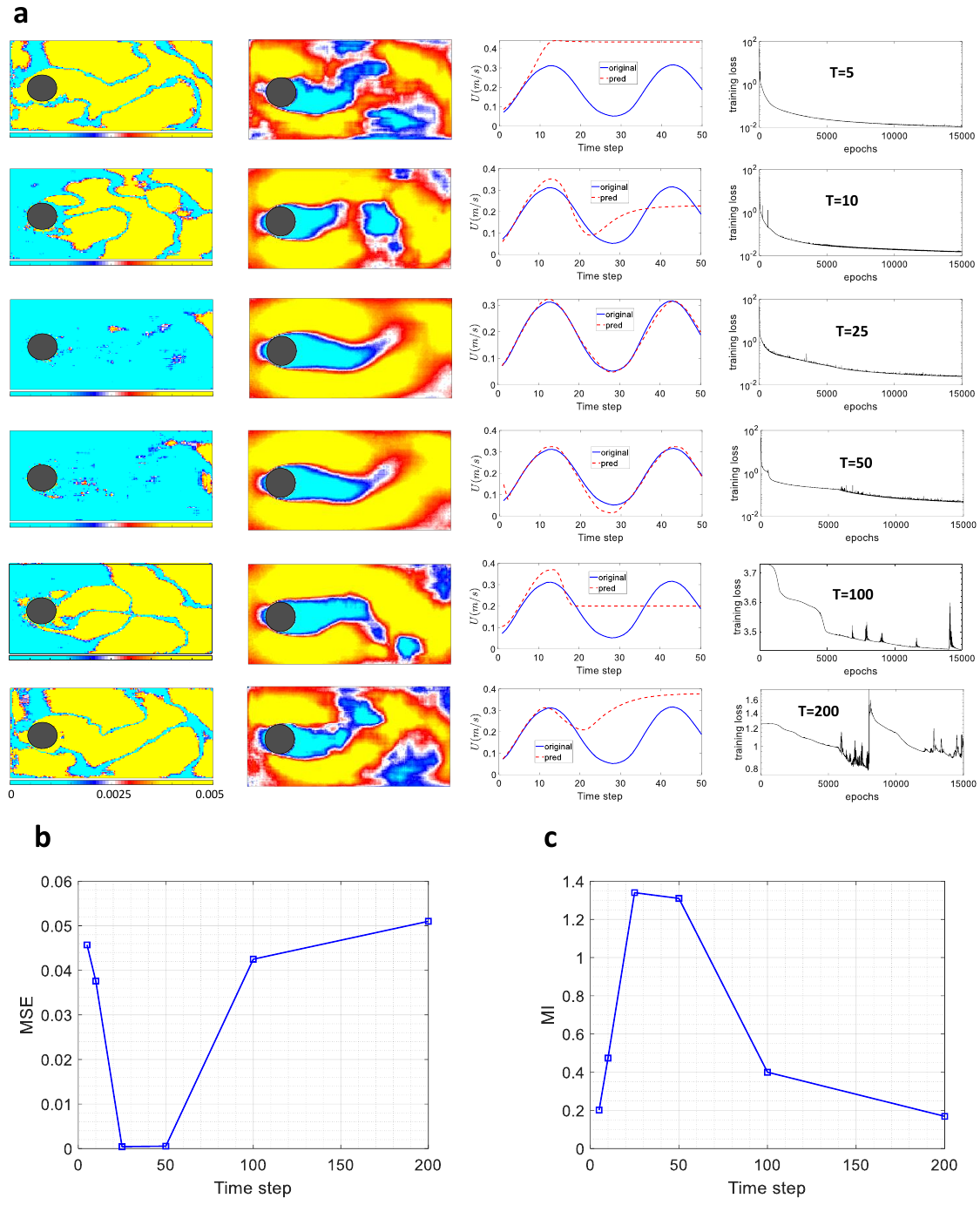}
	\caption{\textbf{a} The impact of the number of time steps on prediction accuracy (in both time series and phase portraits) and the associated training loss curves are analyzed for long-term predictions.
\textbf{b} Assessing the accuracy of predictions using various numbers of time steps with mean squared error~(MSE) and \textbf{c} mutual information~(MI) }
	\label{FIG:11}
\end{figure}

Fig. ~\ref{FIG:11}(a) illustrates the detailed prediction results, including the time series of a selected physical point in the channel, spatial domain predictions, and training loss curves for each training set. We have trained our dynamics generator for a range of 5 to 200 time steps, and we have found that the model is unable to make accurate predictions when it is trained for low or high time steps. It is observed that the training loss curve is smooth when the number of predictions required is small (e.g., $T=5$ or $T=10$), since it is not difficult for the dynamics block to be trained for short-term predictions. However, the model has not been trained to be robust against error accumulation during long-term recursive prediction (here 100 time steps), so the testing prediction over long-term prediction is poor. On the other hand, training loss curves do not stabilize (no convergence) for long prediction horizons (e.g., $T=100$ or $200$ ), leading to poor performance in testing predictions. In the dynamics block, there is an approximate optimal value $T=25$ for prediction horizon, in which training convergence and model robustness are balanced in order to achieve a good long-term prediction of flow fields.

We examine this phenomenon for a stream-wise velocity flow field and find that the most promising results are obtained when 25 time steps are used. Two metrics are used to quantify the effects of each scenario as presented in Fig. \ref{FIG:11}(b) and Fig. \ref{FIG:11}(c). To evaluate the performance of predictions, mutual information (MI) and mean squared error (MSE) are computed between predictions and ground truths~(MSE between ground truths and predicted velocity flow fields and MI between ground truths time series and predicted velocity time series).

In brief, assuming $(X,Y)$ are two random variables with values over $X\times Y$, if their joint distributions is $P_{XY}$ and the marginal distributions are $P_{X}$ and $P_{Y}$, mutual information may be defined as: $I(X;Y)=D_{KL}(P_{(X,Y)}||P_{X} \otimes P_{Y})$ where $D_{KL}$ is Kullback–Leibler divergence~\cite{csiszar1989geometric}. In contrast to Pearson correlation coefficients, which are only capable of detecting linear relationships between variables, mutual information functions can identify nonlinear relationships. We utilize the MI function available in the scikit-learn library in Python. Fig.~\ref{FIG:11}(b) and Fig.~\ref{FIG:11}(c) demonstrate how the number of time steps in the dynamics block affects prediction accuracy. In the case of $T=25$, the prediction exhibits the highest mutual information value and the lowest mean squared error value, which corresponds to approximately the optimal value for training.

\begin{figure}
	\centering
	\includegraphics[width=0.95\textwidth]{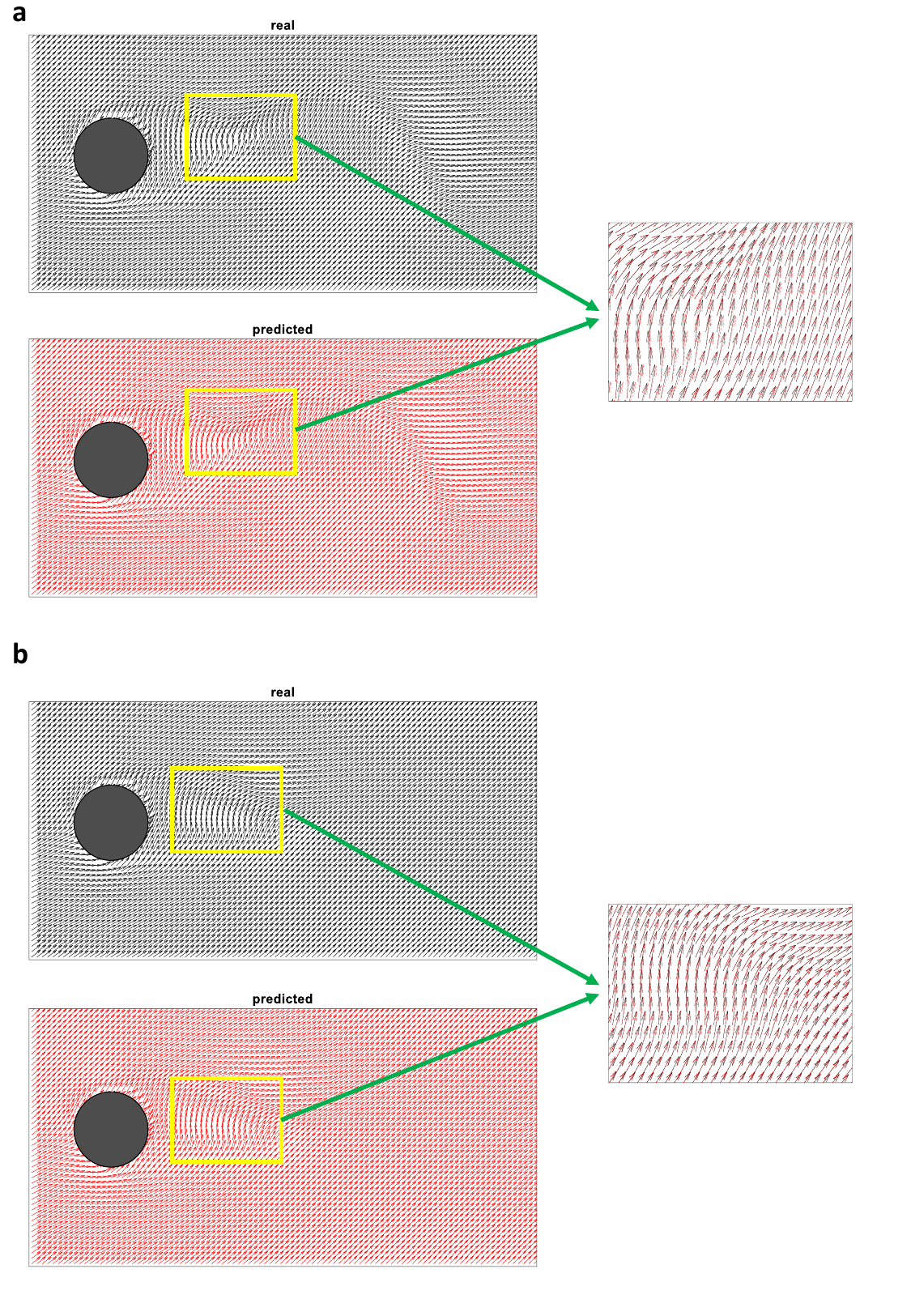}
	\caption{Stream lines of flow passes over a cylinder:\textbf{a} transient stream line showing the predicted velocity in both stream-wise and transverse direction.\textbf{b}  steady-state stream line showing the predicted velocity in both stream-wise and transverse direction.}
	\label{FIG:10}
\end{figure}

\section{Conclusion}
In this study, for data-driven surrogate modeling of fluid dynamics conditioned on certain parameters, we effectively added dynamic constraints to a generative model, predicting flow field sequences for various fluid dynamics systems under both steady-state and transient conditions across a broad range of Reynolds numbers. Results showed that the accuracy of predictions decreases as Reynolds numbers increase, but the framework performs well in both spatial and temporal domains. The generator's predictions are aided by the adversarial loss, and the generator's architecture with an embedded dynamics block effectively captures the underlying flow dynamics.

A number of potential limitations need to be explored further. Firstly, long-term prediction poses a challenge for all data-driven methods, requiring a model that is robust to the errors associated with temporally local predictions. In model training, selecting the right number of time steps (as a hyperparameter) in order to achieve the right balance between convergence at the training phase and the robustness desired is an important part of improving the actual robustness of the model. It is, however, often difficult to determine the optimal hyperparameter. Adaptive training strategies with an adaptive number of time steps may enhance training convergence and, consequently, yield an improvement in model robustness. Second, while the outcome of this study is from typical fluid dynamical systems (such as flow over a cylinder in laminar regime), further validation of the presented method is necessary, especially in the case of turbulent and three-dimensional fluid flows including experimental data from Particle Image Velocimetry (PIV).

\section*{Conflict of interest}
The authors declare no conflict of interest.

\section*{Funding}
This research was partially funded by Federal Highway Administration, the Research Excellence Fund (REF) and faculty startup fund of Michigan Technological University.

\bibliographystyle{jfm}
\bibliography{jfm-instructions}

\end{document}